\documentclass[sigconf,nonacm]{acmart}

\setcopyright{none}

\usepackage{booktabs}
\usepackage{multirow}
\usepackage{graphicx}
\usepackage{subcaption}
\usepackage{xcolor}
\usepackage{balance}
\usepackage{stfloats}
\usepackage{fancyhdr}

\title{Is Grep All You Need? \\How Agent Harnesses Reshape Agentic Search}

\author{Sahil Sen}
\affiliation{\institution{PricewaterhouseCoopers, U.S.}\country{}}
\email{sahil.s.sen@pwc.com}

\author{Akhil Kasturi}
\affiliation{\institution{PricewaterhouseCoopers, U.S.}\country{}}
\email{akhil.kasturi@pwc.com}

\author{Elias Lumer}
\affiliation{\institution{PricewaterhouseCoopers, U.S.}\country{}}
\email{elias.lumer@pwc.com}

\author{Anmol Gulati}
\affiliation{\institution{PricewaterhouseCoopers, U.S.}\country{}}

\author{Vamse Kumar Subbiah}
\affiliation{\institution{PricewaterhouseCoopers, U.S.}\country{}}

\begin{document}


\begin{abstract}
Recent advances in Large Language Model (LLM) agents have enabled complex agentic workflows
where models autonomously retrieve information, call tools, and reason over large corpora to
complete tasks on behalf of users. Despite the growing adoption of retrieval-augmented
generation (RAG) in agentic systems, existing literature lacks a systematic comparison of how
retrieval strategy choice interacts with agent architecture and tool-calling paradigm.
Important practical dimensions, including how tool outputs are presented to the model and how
performance changes when searches must cope with more irrelevant surrounding text, remain
under-explored in agent loops. This paper reports an empirical study organized into two
experiments. Experiment~1 compares grep and vector retrieval on a 116-question sample
from LongMemEval, using a custom agent harness (Chronos) and provider-native CLI harnesses
(Claude Code, Codex, and Gemini CLI), for both inline tool results and file-based tool results
that the model reads separately. Experiment~2 compares grep-only and vector-only retrieval
while progressively mixing in additional unrelated conversation history, so that each query is
embedded in more distracting material alongside the passages that matter. Across Chronos and
the provider CLIs, grep generally yields higher accuracy than vector retrieval in our
comparisons in experiment 1; at the same time, overall scores still depend strongly on which harness and
tool-calling style is used, even when the underlying conversation data are the same.
\end{abstract}


\maketitle

\fancypagestyle{plain}{
  \fancyhf{}
  \fancyfoot[C]{\thepage}
  \renewcommand{\headrulewidth}{0pt}
  \renewcommand{\footrulewidth}{0pt}
}

\fancypagestyle{senstyle}{
  \fancyhf{}
  \fancyhead[R]{Sen et al.}
  \fancyfoot[C]{\thepage}
  \renewcommand{\headrulewidth}{0pt}
  \renewcommand{\footrulewidth}{0pt}
}

\thispagestyle{plain}
\pagestyle{senstyle}


\section{Introduction}
  \label{sec:intro}
  Modern LLM agents increasingly rely on RAG to access external knowledge at inference time~\cite{lewis2020rag,gao2024ragsurvey}, enabling them to reason over corpora that far exceed their context windows. Through tool calling, agents issue search queries, receive ranked results, and iteratively refine their understanding before producing an answer~\cite{yao2023react,schick2023toolformer,qin2023toolsurvey}. Two retrieval paradigms dominate this landscape: semantic vector search, which embeds queries and documents into a shared latent space for approximate nearest-neighbor matching~\cite{karpukhin2020dpr}, and lexical search (e.g., grep, BM25, regex), which performs exact or pattern-based matching over raw text. While vector search has become the default choice in most RAG systems~\cite{gao2024ragsurvey,wang2024ragbestpractices}, lexical search remains widely used in practice due to its simplicity, stability, and low embedding cost~\cite{lin2019neuralhype,thakur2021beir}. However, how retrieval strategy interacts with agent architecture and tool-calling paradigm in end-to-end agentic workflows remains poorly understood.
  Despite growing adoption of agentic search~\cite{asai2024selfrag,jiang2023flare,trivedi2023ircot}, existing research evaluates retrieval strategies largely in isolation from agent architecture. The information retrieval community has extensively benchmarked lexical and dense retrieval
  methods~\cite{thakur2021beir,luan2021sparse,formal2021splade}, and has studied retrieval quality, chunking, and reranking in standalone
  pipelines~\cite{gao2024ragsurvey,wang2024ragbestpractices}. Yet, these evaluations typically assume a fixed pipeline where retrieved documents are concatenated into a prompt, ignoring the iterative, tool-mediated retrieval loops that characterize modern agentic
  systems~\cite{qin2023toolllm,patil2023gorilla}. In practice, agents receive ranked lists but do not treat them as terminal: they decide what to search, how many queries to issue, and whether the retrieved results are sufficient or require further refinement, all mediated by the agent harness and its tool-calling interface~\cite{sumers2023coala,wang2023agentsurvey}. Furthermore, how tool results are
  presented to the model, whether injected inline into the context window or written to files
  that the agent must explicitly read, introduces an additional architectural consideration that
  prior work has not examined.
  At the same time, the emergence of provider-native CLI agents such as Claude Code (Anthropic),
  Codex (OpenAI), and Gemini CLI (Google) has created a new class of agentic systems that differ
  fundamentally from custom-built harnesses~\cite{yang2024sweagent}. These provider harnesses
  embed tool calling into a shell-based interface where the model has direct access to command-line execution tools such as grep, while custom harnesses and agent SDKs offer fine-grained control over the tool-calling
  loop, context construction, and result formatting. How retrieval strategy effectiveness varies
  across these architecturally distinct harness classes has not been studied. Additionally, a
  less-examined dimension of retrieval quality is robustness to corpus noise: as the ratio of
  irrelevant to relevant documents increases, retrieval strategies may degrade at different
  rates~\cite{liu2023lostinmiddle}, and understanding this scaling behavior is critical for
  deploying RAG systems against large, noisy corpora.
  This paper aims to close these gaps with an empirical study of retrieval strategies for
  tool-equipped LLM agents, organized by experiment (Section~\ref{sec:experiments}). We evaluate
   lexical search and semantic vector search across a custom harness (Chronos~\cite{chronos}) and
   provider-native CLI harnesses (Claude Code, Codex, Gemini CLI), under both standard inline
  context delivery and programmatic file-based result delivery. We evaluate multiple LLMs on a
  116-question subset of the LongMemEval benchmark~\cite{longmemeval} spanning six categories of
  information retrieval tasks. The paper contributes in three ways:
  \begin{itemize}
    \item \textbf{Retrieval, harness, and presentation}. Evidence on how the choice between
  lexical and dense retrieval combines with the agent orchestration layer and with whether tool
  outputs are surfaced inline or through files.
    \item \textbf{Noise and scale}. Characterization of how end-to-end behavior evolves as
  irrelevant surrounding content grows relative to task-relevant material, including interactions
   between retriever behavior and the broader agent loop.
    \item \textbf{Heterogeneity across agent stacks}. A direct comparison showing that retrieval
  effectiveness is not stable across architecturally distinct harnesses (custom versus
  provider-native CLIs) even when the underlying text corpus is held fixed.
  \end{itemize}


\section{Overview of Retrieval in Agentic Systems}
\label{sec:overview}

Retrieval in agentic systems refers to the process by which an LLM agent identifies, executes, and consumes search operations over a corpus to answer a user query. Unlike standalone retrieval pipelines, where a fixed query is matched against a document index and the top-$k$ results are concatenated into a prompt~\cite{lewis2020rag,gao2024ragsurvey}, agentic retrieval is iterative and agent-directed: the model decides what to search for, how many queries to issue, and whether the retrieved results are sufficient or require refinement~\cite{yao2023react,jiang2023flare,asai2024selfrag}. This process is mediated by two design dimensions that jointly determine end-to-end effectiveness: the retrieval strategy (lexical, semantic, or hybrid), and the agent harness (custom or provider-native).

\subsection{Retrieval Strategies}
\label{sec:retrieval-strategies}

Retrieval strategies for agentic search systems fall into three broad categories: lexical, semantic, and hybrid. Each offers distinct tradeoffs in accuracy, latency, cost, and robustness to query formulation.

\subsubsection{Lexical Search.}
Lexical retrieval methods perform exact or pattern-based matching over raw text. Classical approaches such as BM25~\cite{lin2019neuralhype} score documents by term frequency and inverse document frequency, while grep search uses regular expressions or substring matching to locate passages containing specific keywords~\cite{lumer2025toolsurvey}. Lexical methods require no embedding model or vector index, and incur negligible computational cost beyond the text scan itself. The BEIR benchmark demonstrated that BM25 remains a competitive baseline across diverse retrieval tasks, often outperforming early dense retrieval models in zero-shot settings~\cite{thakur2021beir}. Learned sparse representations such as SPLADE~\cite{formal2021splade} extend lexical matching by expanding query and document terms through a learned vocabulary, bridging the gap between exact matching and semantic understanding while preserving the interpretability and efficiency of sparse representations.

\subsubsection{Semantic Search.}
Semantic or dense retrieval encodes queries and documents as dense vectors in a shared embedding space and retrieves nearest neighbors, with the most common way being approximate nearest-neighbor (ANN) search~\cite{karpukhin2020dpr}. Dense passage retrieval (DPR) established this paradigm by training dual encoders on question-passage pairs, enabling retrieval based on meaning rather than surface-level term overlap~\cite{karpukhin2020dpr}. RAG systems extended this by coupling dense retrieval with a generative model, allowing the retriever and generator to be jointly optimized~\cite{lewis2020rag}. Modern RAG pipelines typically use pre-trained embedding models to encode documents at indexing time and queries at inference time, with optional post-retrieval reranking to refine the initial candidate set~\cite{gao2024ragsurvey,wang2024ragbestpractices}. While semantic search excels at handling paraphrases and semantic similarity, it introduces dependencies on embedding model quality, vector index infrastructure, and indexing latency that lexical methods avoid.

\subsubsection{Hybrid Approaches.}
Hybrid retrieval combines lexical and semantic signals to leverage the strengths of both paradigms. Reciprocal rank fusion (RRF)~\cite{cormack2009rrf,frtr2026} merges ranked lists from independent lexical and dense retrievers without requiring score calibration. Late interaction models such as ColBERT~\cite{khattab2020colbert} compute fine-grained token-level similarity between query and document representations, achieving a middle ground between the efficiency of single-vector retrieval and the expressiveness of cross-encoder reranking. Investigations into the complementarity of sparse and dense representations have shown that lexical and semantic methods often retrieve different relevant documents, making their combination more effective than either alone~\cite{luan2021sparse}. In agentic settings, hybrid retrieval can also emerge organically when agents have access to both lexical and semantic search tools and choose between them based on the query.

\subsection{Agent Harnesses}
\label{sec:harnesses}
The agent harness is the environment layer that manages the tool-calling loop: it constructs the prompt, dispatches tool calls, receives results, and decides whether to continue iterating or produce a final answer. We distinguish between two classes of harnesses that differ fundamentally in the degree of control they afford over this process.

\subsubsection{Custom Harnesses.}
Custom harnesses are built by developers using agent frameworks, provider open SDKs, or custom code~\cite{yao2023react,sumers2023coala}. These provide fine-grained control over every stage of the agentic loop: the system prompt, tool definitions, context construction, result formatting, and iteration termination criteria. The ReAct paradigm~\cite{yao2023react}, which interleaves reasoning traces with tool actions, is the most widely adopted pattern for custom harnesses. Developers can implement domain-specific optimizations such as dynamic prompting (tailoring the system prompt per query), result truncation policies, and reranking of retrieved passages. Custom harnesses also allow explicit management of the context window, for example by summarizing or discarding earlier tool results as the conversation grows~\cite{memtool2025,packer2023memgpt}. The tradeoff is developmental overhead: building and maintaining a custom harness requires significant engineering effort and expertise in prompt engineering, tool interface design, and context management.

\subsubsection{Provider-Native CLI Harnesses.}
Provider-native CLI harnesses embed tool calling into a shell-based interface where the model has direct access to system utilities~\cite{yang2024sweagent,chen2021codex}. In these environments, the agent can execute arbitrary bash commands, including grep, find, cat, and other Unix tools, as native tool actions. The harness manages context construction, and iteration control according to the provider's internal implementation, which is largely opaque to the user. Provider-native harnesses offer minimal setup cost and leverage the provider's optimized context engineering, but they sacrifice the fine-grained control available in custom harnesses. Notably, when grep is available as a native bash tool, the boundary between ``retrieval strategy'' and ``agent capability'' blurs: the agent can construct its own grep commands, choosing search terms, flags, and file targets dynamically, rather than being limited to a predefined search API.

\subsection{Tool-Calling Architectures}
\label{sec:tool-calling}
Orthogonal to the choice of harness, tool-calling architectures govern how retrieval results are delivered to the model after a search is executed. This design dimension has significant implications for context window utilization and the agent's ability to process large result sets. tool calling method mode is, in effect, a context engineering decision: inline delivery fills the context window directly, while programmatic delivery delegates context construction to the agent itself.

\subsubsection{Standard (Inline).}
In standard tool-calling architectures, search results are returned directly as tool response messages appended to the conversation context~\cite{schick2023toolformer,qin2023toolllm,patil2023gorilla}. The model receives the full result set within the context window and can immediately reason over it. This is the default mode for native function calling in most LLM APIs and custom harnesses. The main advantage is simplicity: retrieval results appear inline with the dialogue, without a separate buffer, routing layer, or post processing step between fetching and generation. However, large result sets compete for context window space with the system prompt, conversation history, and previous tool results, creating context pressure --- a phenomenon sometimes called \emph{context rot} --- that can degrade performance on long-horizon tasks~\cite{liu2023lostinmiddle}. One method of mitigating this is by applying result truncation at the cost of potentially discarding relevant information.

\subsubsection{Programmatic (File-Based).}
In programmatic tool-calling architectures, search results are written to disk and the model receives only a file path or summary pointer~\cite{packer2023memgpt,lumer2025toolsurvey}. The agent must then take an explicit action to access the results, which can itself be a search operation (e.g. \texttt{grep} over the result file) or a full read (e.g. \texttt{cat}, \texttt{read\_file}). This decouples retrieval result size from context window pressure: arbitrarily large result sets can be written to disk without consuming context tokens until the agent explicitly reads them. The tradeoff is indirectness: the agent must execute an additional tool call to access results, which adds latency and requires the model to understand the file-based workflow. Programmatic architectures also enable progressive disclosure, where the agent reads only a subset of results based on metadata or summaries, a form of agent-driven post-retrieval filtering that is not possible with inline delivery.


\section{Methodology}
\label{sec:method}
Our experiments evaluate retrieval strategy effectiveness across the two dimensions defined in Section~\ref{sec:overview}.

\subsection{Task and Dataset}
\label{sec:dataset}

We evaluate on a 116-question representative subset of the LongMemEval benchmark~\cite{longmemeval}, which tests an agent's ability to answer questions over long conversations spanning multiple sessions. Each question is accompanied by sessions of a certain type: one or more oracle sessions containing the information needed to answer correctly, and a variable number of distractor sessions that are irrelevant to the query. Questions span six categories: knowledge-update (tracking state changes over time), multi-session (aggregating information across sessions), single-session-assistant (recalling model-generated content), single-session-preference (user personal preferences), single-session-user (user-stated facts), and temporal-reasoning (computing durations, ordering events, and resolving dates). All conversation turns and extracted temporal events are stored locally, serving as the corpus for both grep and vector search.

\subsection{Retrieval Implementations}
\label{sec:retrieval}

\subsubsection{Structured events via Chronos.}
Our search layer operates over per-question files, serializing LongMemEval dialogue turns together with structured temporal events extracted from transcripts using the Chronos preprocessing pipeline~\cite{chronos}. Chronos targets long-horizon conversation memory by surfacing salient time structure, including explicit dates, intervals, and related time spans, as first-class text records that are coupled with raw utterances, consistent with Chronos's emphasis on structured event retrieval for temporal-aware agents~\cite{chronos}. We adopt this layer for two reasons:
\begin{itemize}
    \item Our study evaluates the agent's search techniques, not its temporal reasoning. A substantial fraction of LongMemEval items depend on explicit time expressions scattered across many turns; surfacing these as normalized records in a compact parallel channel ensures that success on temporal items reflects whether the agent can locate the relevant evidence, rather than whether the model can reconstruct dates and intervals from fragmented mentions.
    \item Chronos's structured event extraction mirrors the preprocessing a long-memory agent would use in deployment, so the experimental setup reflects a realistic production configuration rather than an artificial one constructed only for evaluation.
\end{itemize}

\subsubsection{Lexical Search (Grep).}
The grep retrieval tool loads conversation turns and extracted temporal events from per-question files into memory and executes regular expression (regex) matching over the raw text fields. Results are scored by match count and returned. The implementation requires no embedding model, vector index, or external service: all matching is performed in-process over local files.

\subsubsection{Semantic Search (Vector).}
  The vector search tool queries a search index populated at ingestion time. Each conversation turn and temporal event is embedded and stored in a per-question index with a versioned schema. At query
  time, the tool embeds the natural language query and retrieves the most relevant results using approximate nearest-neighbor search. A reranking
  step re-scores retrieved passages before returning the top-$k$ results, with $k$ being selected by the agent, to the agent~\cite{gao2024ragsurvey}.

\subsection{Agent Harnesses}
\label{sec:method-harnesses}

\subsubsection{Custom Harness.}
Our custom harness, Chronos, implements an agent using LangChain with access to four search tools (grep and vector search over turns and events). In addition, a grep tool is enabled for programmatic mode. These tools are enabled or disabled based on the experimental configuration. Following Chronos's design for long-horizon memory agents~\cite{chronos}, we use dynamic prompting: the system instructions, search hints, and tool-use guidance depend on the detected question category (e.g., temporal reasoning versus preference recall), rather than a single static system prompt for all items. The agent therefore starts each episode with category-conditioned guidance, followed by an initial broad context block (top-15 vector results) before the tool-calling loop begins. The loop continues until the model produces a final answer.

 \subsubsection{Provider-Native CLI Harnesses.}
  We evaluate three provider-native CLI agents~\cite{yang2024sweagent}. Each receives the question and a dynamically generated search strategy, and can invoke bash-callable wrapper scripts for grep and
  vector search via absolute paths. In standard mode, the process was spawned in a sandbox to ensure the model has access to only relevant files.

\subsection{Tool-Calling Architectures}
\label{sec:toolcalling}

\subsubsection{Standard (Inline).}
In standard mode, search scripts print results directly to stdout. For the Chronos harness, results are returned as tool response messages injected into the conversation context. For CLI harnesses, stdout is appended to the agent's working context by the shell interface. Results compete with the system prompt and conversation history for context window space.

\subsubsection{Programmatic (File-Based).}
  In programmatic mode, all results are written to a file, to which the agent has access to. The file is then read or searched through by the agent to access the full results. This decouples retrieval result size from context
  pressure and enforces that the agent explicitly select which results to consume~\cite{packer2023memgpt}.

\subsection{Models}
\label{sec:models}

We evaluate five LLMs and a range of capability levels: Claude Opus 4.6 and Claude Haiku 4.5, GPT-5.4, and Gemini 3.1 Pro and Gemini 3.1 Flash-Lite.

\subsection{Evaluation}
  \label{sec:eval}
  Following the evaluation protocol specified in the LongMemEval benchmark paper~\cite{longmemeval}, we assess each model hypothesis with an auxiliary LLM grader. We instantiated GPT-4o as the metric model: for every question, the grader receives the question text, the reference answer field, and the agent's hypothesis, and must
  output a binary judgment under category-conditioned instructions (e.g., tolerance for off-by-one temporal counts, rubric-style scoring for preference items, and abstention handling for \texttt{\_abs}
  variants). We report accuracy as the fraction of questions for which the grader answers affirmatively. Holding the grader model, prompt templates, and decoding settings fixed across conditions ensures
  that differences across harnesses, retrieval modes, and session-limit settings reflect changes in the agent pipeline rather than evaluation noise.


\section{Experiments}
\label{sec:experiments}

\subsection{Experiment 1: Retrieval Mode, Harness, and Tool Calling Method}
\label{sec:exp1}

 \subsubsection{Goal}
  \label{sec:exp1-goal}
  We isolate how retrieval mode (grep-only vs.\ vector-only), agent harness (Chronos vs.\ Claude Code vs.\ Codex vs.\ Gemini CLI), and tool calling method (standard inline vs.\ programmatic file-based)
  jointly affect end-to-end long-memory QA accuracy when the full per-question haystack is exposed. The factorial design answers whether lexical or dense retrieval has a consistent advantage under matched
   harnesses, and whether routing tool outputs to files instead of inline messages changes that comparison.

\subsubsection{Experimental setup}
\label{sec:exp1-setup}

We use the 116-question LongMemEval-S subset, corpus, grep and vector tool implementations, harness configurations, models, and GPT-4o grader described in Section~\ref{sec:method}. Each row of Table~\ref{tab:exp1-full} fixes one harness--model pair and varies retrieval mode and tool calling method as defined in Sections~\ref{sec:retrieval}--\ref{sec:toolcalling}.

\subsubsection{Results}
  \label{sec:exp1-results}
  Table~\ref{tab:exp1-full} reports overall accuracy (\%) on the 116-question LongMemEval-S
  subset~\cite{longmemeval} over Chronos-processed session JSON~\cite{chronos} with the full
  per-question haystack.
  Columns cross grep-only versus vector-only retrieval with inline versus programmatic tool
  delivery (Section~\ref{sec:toolcalling}).
  With inline delivery, lexical search is uniformly stronger than dense retrieval: inline grep
  exceeds inline vector for every harness--model pair.
  The largest margin is Chronos with Gemini~3.1~Flash-Lite (86.2\% vs.\ 62.9\%), and the
  narrowest is Claude Code with Claude Opus~4.6 (76.7\% vs.\ 75.0\%).
  On Chronos, inline grep spans 83.6--93.1\% across backbones, whereas inline vector spans
  62.9--83.6\%.
  The same Claude Opus~4.6 backbone reaches 93.1\% under Chronos but 76.7\% under Claude Code, so
   changing the harness shifts the ceiling by roughly as much as swapping retrievers within a
  fixed harness.
  Codex with GPT-5.4 ties the strongest Chronos inline grep (93.1\%) while its inline vector
  accuracy is 75.9\%.
  Programmatic delivery reshuffles the comparison: programmatic vector exceeds programmatic grep
  on five of ten harness--model pairs (Chronos with Claude Opus~4.6; Claude Code with Claude
  Opus~4.6; Codex with GPT-5.4; Gemini CLI with Gemini~3.1~Flash-Lite; Gemini CLI with
  Gemini~3.1~Pro), while programmatic grep remains higher on the other Chronos backbones and on
  Claude Code with Claude Haiku~4.5.
  The sharpest regression is Codex with GPT-5.4, which falls from 93.1\% under inline grep to
  55.2\% under programmatic grep, with programmatic vector at 67.2\% for that same pair.
  Other notable spreads include Claude Haiku~4.5 on Claude Code (55.2\% inline grep vs.\ 44.0\% inline vector, and 37.1\% vs.\ 32.8\% programmatic) and Gemini~3.1~Flash-Lite on Gemini CLI (87.1\% vs.\ 67.2\% inline; 68.1\% vs.\ 74.1\% programmatic, one of the few cases where vector leads under programmatic delivery on that harness).
  The full numeric grid is Table~\ref{tab:exp1-full}.

\begin{table*}[t]
        \caption{Experiment 1. Overall accuracy (\%) on the 116-question LongMemEval-S
    subset. We report results for both the standard and programmatic tool-calling configurations}
        \label{tab:exp1-full}
        \centering
        \small
        \begin{tabular}{llcccc}
          \toprule
          Model & Harness & grep & vector &
    \shortstack{grep\\programmatic} &
    \shortstack{vector\\programmatic} \\
          \midrule
          Claude Opus 4.6 & Chronos      & 93.1 & 83.6 & 80.2 & 81.9 \\
          Claude Opus 4.6 & Claude Code  & 76.7 & 75.0 & 68.1 & 79.3 \\
          \midrule
          Claude Haiku 4.5 & Chronos      & 83.6 & 76.7 & 83.6 & 81.9 \\
          Claude Haiku 4.5 & Claude Code  & 55.2 & 44.0 & 37.1 & 32.8 \\
          \midrule
          GPT-5.4 & Chronos      & 89.7 & 81.9 & 87.1 & 75.0 \\
          GPT-5.4 & Codex CLI    & 93.1 & 75.9 & 55.2 & 67.2 \\
          \midrule
          Gemini 3.1 Pro & Chronos      & 91.4 & 82.8 & 79.3 & 76.7 \\
          Gemini 3.1 Pro & Gemini CLI   & 81.9 & 75.0 & 81.0 & 82.8 \\
          \midrule
          Gemini 3.1 Flash-Lite & Chronos      & 86.2 & 62.9 & 85.3 & 72.4 \\
          Gemini 3.1 Flash-Lite & Gemini CLI   & 87.1 & 67.2 & 68.1 & 74.1 \\
          \bottomrule
        \end{tabular}
    \end{table*}

  Per-category accuracy for the Chronos harness (grep-only, programmatic tool calling method, full haystack) appears in Appendix~\ref{app:per-category} (Table~\ref{tab:category}).

\subsubsection{Discussion}
\label{sec:exp1-discussion}

Taken together, the factorial layout suggests two interpretive points. First, LongMemEval rewards recovering literal witnesses: exact dates, counts, preferences, and spans that often remain stable under tokenization. Lexical tools surface those strings without an embedding bottleneck, which could explain why  inline grep is a strong default in table~\ref{tab:exp1-full}. A per-category breakdown of the results can be found in Appendix~\ref{app:per-category}. Second, ``retrieval mode`` is not measured in isolation: the harness shapes the system prompt, tool descriptions, and how hits are rendered back into the chat, all of which influence how the model schedules queries and decides when to stop. The same underlying model therefore operates under very different framing across harnesses, which plausibly contributes to the harness-level shifts observed in Table~\ref{tab:exp1-full}~\cite{yao2023react,sumers2023coala}.

Programmatic delivery changes the task from ``read the tool message'' to ``locate, open, and integrate an artifact''~\cite{packer2023memgpt}. When that second stage is brittle, accuracy can collapse independently of retrieval quality, which helps explain why programmatic conditions reorder grep--vector comparisons without any change to the index. A practical implication is that agent papers should report both retrieval mechanics and the delivery path, because file-based routing is itself a tool-use stress test.

Lexical and dense search optimize different failure modes in an agent loop, not only in a ranking metric. Grep is deliberately narrow: it rewards the model for generating high-precision patterns, but it punishes vocabulary mismatch, if the agent never guesses a distinctive substring, nothing is retrieved. Dense retrieval is deliberately broad: it can surface paraphrases and oblique mentions, but it also elevates semantically ``near'' distractors that share topic overlap with the question, which matters when the agent issues short or underspecified queries. Table~\ref{tab:exp1-full} is consistent with a regime where LongMemEval's answers are often licensed by a small set of literal spans, so the precision bias of lexical matching wins when hits are injected inline and immediately actionable.

Moving the same backbone (e.g., Claude Opus~4.6) between Chronos and a provider CLI changes accuracy by margins comparable to swapping retrievers inside a fixed harness (Section~\ref{sec:exp1-results}). One way to read this is that a harness is not passive infrastructure: Chronos's category-conditioned prompting~\cite{chronos} and controlled tool surface area shape what gets searched first and how failures are repaired, whereas CLI agents inherit provider-specific tool ergonomics, sandboxing, and transcript formatting. In other words, ``retrieval'' in Table~\ref{tab:exp1-full} is really retrieval-plus-orchestration. This is a constructive takeaway for benchmarks: reporting only BM25 vs.\ ANN in a static pipeline under-estimates the variance introduced by agent scaffolding.

File-based delivery is often motivated as relief from context pressure, which should disproportionately benefit settings where inline vector dumps crowd the window~\cite{packer2023memgpt,liu2023lostinmiddle}. We indeed see programmatic vector exceed programmatic grep on several rows. Yet the same mechanism can invert: if the model struggles to complete the read--integrate--retry cycle, the benefit never reaches the answer layer. The extreme Codex/GPT-5.4 programmatic grep regression in Section~\ref{sec:exp1-results} is a useful cautionary tale: cheap retrieval (regex over local JSON) is not ``easy'' end-to-end if the harness turns each hit into a multi-step workflow that the stack executes unreliably. This suggests a design tradeoff: programmatic routing trades context bandwidth for compositional tool competence; gains are realized only when the agent reliably closes the loop.

The Claude Haiku~4.5 rows on Claude Code show especially large inline grep--vector gaps. Without trace-level causal attribution, a plausible hypothesis is that weaker models are less consistent at iterative query refinement and reranker-aware reading, which hurts dense retrieval more than pattern-triggered lexical recovery when evidence is literally present. If true, ``default to vector'' recommendations should be conditioned on backbone strength and on whether the task rewards literal span recovery versus conceptual blending: a nuance that aggregate leaderboard comparisons often obscure.

\subsection{Experiment 2: Context Scaling with Increasing Noise}
\label{sec:exp2}

\subsubsection{Goal}
\label{sec:exp2-goal}
Experiment~2 asks how lexical and dense retrieval diverge as the model is exposed to more sessions from the same per-question bundle, rather than only documenting that accuracy drops when more irrelevant dialogue is present.
Long-memory QA interleaves oracle sessions with distractors; as the session limit increases under a fixed sampling protocol, both retrievers see additional irrelevant material.
We sweep a discrete grid of such settings (Section~\ref{sec:scaling}), holding tool-delivery conventions fixed, and report paired grep-only and vector-only tables so each column reflects identical distractor exposure for the two retrieval families rather than a single full-haystack snapshot.

\subsubsection{Experimental setup}
\label{sec:exp2-setup}
\label{sec:scaling}
To stress-test robustness as irrelevant sessions accumulate around oracle evidence, we vary the per-question session limit, instantiating configurations labeled s5, s10, s20, s30, and full in Tables~\ref{tab:exp2-grep-all}--\ref{tab:exp2-vector-all}, where full denotes the complete haystack (39--66 sessions per item). Oracle sessions required to answer are always retained; remaining slots up to the active limit are filled with distractors sampled from other sessions in the same per-question bundle. We use the task, models, harnesses, and tools from Section~\ref{sec:method}. We report results for both standard grep and standard vector search over each model and harness evaluated in Experiment 1.

\begin{figure*}[t]
    \centering
    \includegraphics[width=\textwidth]{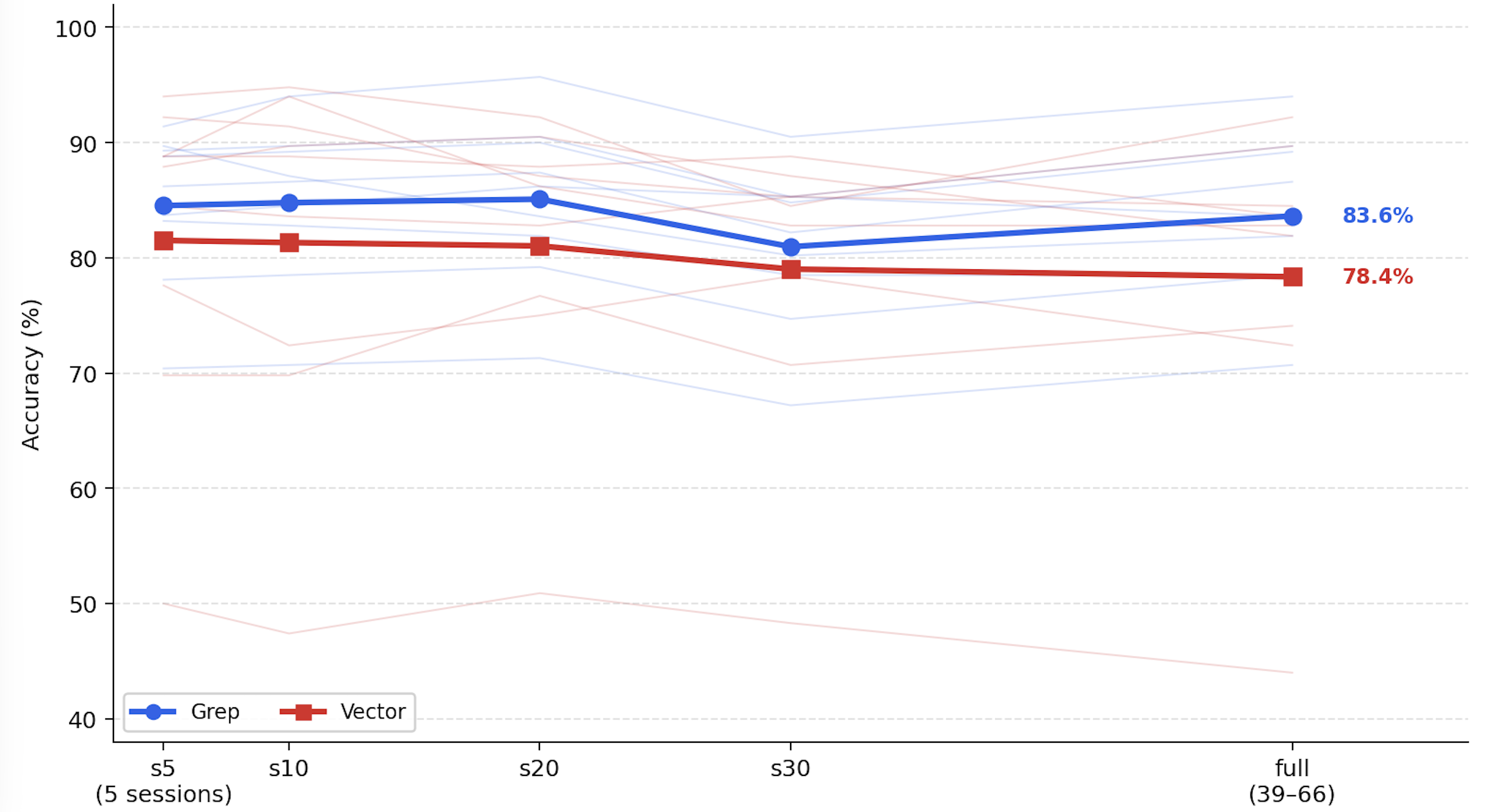}
    \caption{Mean performance as noise is added to the retrieval pool. Both methods face minimal degradation, but GREP outperforms vector on average.}
    \label{fig:fig1}
\end{figure*}

\subsubsection{Results}
\label{sec:exp2-results}
Tables~\ref{tab:exp2-grep-all} and~\ref{tab:exp2-vector-all} pair grep-only and vector-only overall accuracy (\%) on the 116-question subset across session-limit configurations (Section~\ref{sec:scaling}).
Chronos uses grep-only or vector-only retrieval harnesses; Claude Code and Gemini CLI use bash search on the same corpus.
The Codex vector row is complete only at~full; intermediate configurations are pending.

Grep accuracy is not monotone: Chronos Opus rises to 90.5\% at~s20, dips to 85.3\% at~s30, then reaches 89.7\% at~full; Claude Code Opus peaks at~s20 (95.7\%) but finishes at 94.0\% at~full; Gemini CLI Pro is lowest at~s30 (74.7\%).
At~s5, Chronos grep spans 83.2--89.7\% across backbones, while Chronos vector at the same configuration spans 87.9--94.0\%.
Vector trajectories peak at different configurations: Chronos Opus is highest at~s10 (94.8\%), lowest at~s30 (84.5\%), and 92.2\% at~full; Claude Code Opus is weakest at~s10 (72.4\%) and best at~s30 (78.4\%).
Retriever ordering depends on harness: Claude Code favors grep for Opus and Haiku at every configuration we report, Gemini CLI Pro favors vector throughout, and Chronos shows crossings as more sessions are included (e.g., Opus is vector-ahead from s5--s20 and at~full, but grep is ahead at the penultimate limits: 85.3\% vs.\ 84.5\%).
At~full on Chronos, Gemini~3.1~Pro is an example of a late reversal: grep reaches 86.6\% versus 84.5\% for vector despite vector being competitive or stronger at s5--s20. Under Gemini CLI, by contrast, Gemini~3.1~Pro stays vector-ahead at every configuration, with the grep--vector gap widening to 89.7\% vs.\ 78.5\% at~full.
For Flash-Lite, Chronos shows mixed crossings between grep and vector across the grid, whereas the two Gemini CLI Flash-Lite trajectories remain comparatively flat relative to Opus.
The complete grids appear in Tables~\ref{tab:exp2-grep-all} and~\ref{tab:exp2-vector-all}.

\begin{table*}[t]
    \caption{Experiment 2 (grep-only): overall accuracy (\%) on the 116-question subset by session-limit configuration.}
    \label{tab:exp2-grep-all}
    \centering
    \small
    \begin{tabular}{@{}llccccc@{}}
      \toprule
      Model & Harness & s5 & s10 & s20 & s30 & full \\
      \midrule
      Claude Opus 4.6       & Chronos      & 89.3 & 89.7 & 90.5 & 85.3 & 89.7 \\
                            & Claude Code  & 91.4 & 94.0 & 95.7 & 90.5 & 94.0 \\
      \midrule
      Claude Haiku 4.5      & Chronos      & 83.7 & 84.5 & 86.2 & 85.3 & 83.6 \\
                            & Claude Code  & 89.7 & 87.1 & 83.6 & 80.2 & 81.9 \\
      \midrule
      GPT-5.4               & Chronos      & 83.2 & 82.8 & 81.9 & 78.5 & 78.5 \\
      \midrule
      Gemini 3.1 Pro        & Chronos      & 86.2 & 86.6 & 87.4 & 82.2 & 86.6 \\
                            & Gemini CLI   & 78.1 & 78.5 & 79.2 & 74.7 & 78.5 \\
      \midrule
      Gemini 3.1 Flash-Lite & Chronos      & 88.8 & 89.2 & 90.0 & 84.8 & 89.2 \\
                            & Gemini CLI   & 70.4 & 70.7 & 71.3 & 67.2 & 70.7 \\
      \bottomrule
    \end{tabular}
\end{table*}

\begin{table*}[t]
      \caption{Experiment 2 (vector-only): accuracy (\%) by session-limit configuration.}
      \label{tab:exp2-vector-all}
      \centering
      \small
      \setlength{\tabcolsep}{3.5pt}
      \begin{tabular}{@{}llccccc@{}}
        \toprule
        Model & Harness & s5 & s10 & s20 & s30 & full \\
        \midrule
        Claude Opus 4.6       & Chronos      & 94.0 & 94.8 & 92.2 & 84.5 & 92.2 \\
                              & Claude Code  & 77.6 & 72.4 & 75.0 & 78.4 & 72.4 \\
        \midrule
        Claude Haiku 4.5      & Chronos      & 87.9 & 89.7 & 90.5 & 87.1 & 87.9 \\
                              & Claude Code  & 50.0 & 47.4 & 50.9 & 48.3 & 44.0 \\
        \midrule
        GPT-5.4               & Chronos      & 88.8 & 94.0 & 86.2 & 82.8 & 82.8 \\
        \midrule
        Gemini 3.1 Pro        & Chronos      & 92.2 & 91.4 & 87.1 & 85.3 & 84.5 \\
                              & Gemini CLI   & 84.5 & 83.6 & 82.8 & 85.3 & 89.7 \\
        \midrule
        Gemini 3.1 Flash-Lite & Chronos      & 88.8 & 88.8 & 87.9 & 88.8 & 83.6 \\
                              & Gemini CLI   & 69.8 & 69.8 & 76.7 & 70.7 & 74.1 \\
        \bottomrule
      \end{tabular}
  \end{table*}

\subsubsection{Discussion}
\label{sec:exp2-discussion}
A common practitioner intuition holds that lexical search suffices on small corpora but breaks at scale, while more expressive semantic search becomes necessary as corpus size grows. Our scaling study partially supports this --- vector retrieval is often stronger at low session counts --- but reveals that the crossover depends on harness and backbone rather than corpus size alone. The scaling study is therefore most useful as a stress test of interaction: retrieval families do not degrade in parallel as ``noise increases''; they interact with (i) how distractors are resampled for each session-limit configuration, (ii) harness-specific tool transcripts, and (iii) the model's implicit policy for when to stop searching.

A useful abstraction is that dense retrieval tends to explore neighborhoods in embedding space: it can recover indirect mentions, but it also admits topical false friends as sessions accumulate. Lexical retrieval tends to exploit surface cues: it is brittle to phrasing, yet when the agent discovers a discriminative pattern it can be ruthlessly precise. The Chronos scaling grid, where vector is often stronger at smaller session limits but grep can close or overtake later depending on backbone and column, is consistent with a story where semantic retrieval buys early coverage when the bundle is still manageable, while regex-style evidence becomes comparatively stable once the agent must separate needle-from-haystack under a workflow that already consumes transcript tokens for tool bookkeeping. We state this as a hypothesis: our tables do not isolate query strings or retrieved sets, but the qualitative crossings are hard to explain if ``noise'' were a one-dimensional monotone.

The persistent grep advantage on Claude Code for Opus/Haiku and the persistent vector advantage for Gemini~3.1~Pro on Gemini CLI (Section~\ref{sec:exp2-results}) suggest stable inductive biases introduced by provider tooling: default hints, how stdout is chunked into the transcript, tool error surfaces, and even cultural defaults in how the CLI agent phrases searches. This is a reminder that ``grep'' in production is rarely a single primitive, it is grep-plus-shell-plus-prompting. Vendor-stable patterns are scientifically interesting because they imply that migration between CLI stacks is not retrieval-interchangeable even when the on-disk corpus is byte-identical.

Since distractors are redrawn when the session limit changes, mid-grid peaks need not indicate that ``30 sessions is easier than 20'' in any absolute sense; they can indicate favorable interference between a particular sampled distractor set and the agent's search trace~\cite{packer2023memgpt}. That does not diminish the paired comparison: grep and vector still face the same sampled bundle for each configuration. It does mean scaling curves should be read as samples from a stochastic outer loop, not as smooth capacity laws.

Finally, incomplete rows (Codex vector intermediates; no Codex grep scaling row yet) mean we cannot yet state a vendor-complete picture of how ``CLI grep'' ages with distraction relative to ``CLI vector'' under matched caps. Until those rows exist, the strongest cross-experiment claim is conditional: Section~\ref{sec:exp1} establishes large end-to-end differences at full haystack under heterogeneous harnesses; Section~\ref{sec:exp2} shows that the grep--vector ordering is not preserved under scaling even when distractor sampling is paired, but the shape of that non-preservation still depends on which stack is observed.

  \section{Limitations}
  \label{sec:limitations}
  Conceptually, our conclusions are tied to long-memory conversational QA: questions are grounded in multi-session chat, explicit time expressions, and personal/user facts. Lexical tools may be disproportionately helpful here because answers often license on verbatim spans; in domains where evidence is rarely literal (e.g., scientific synthesis over paraphrased abstracts, visual-heavy documents, or code semantics), dense retrieval and hybrid routing may look different. We do not claim that grep ``beats'' vector in general, only that it can win end-to-end under the task distribution and corpora we study.


\section{Conclusion}
\label{sec:conclusion}
As LLM agents take on increasingly complex agentic workflows that autonomously retrieve information, call tools, and reason over large corpora, the choice of retrieval strategy interacts with agent architecture and tool-calling paradigm in ways that existing literature has not systematically compared. Practical dimensions such as how tool outputs are presented to the model, and how performance changes when searches must cope with more irrelevant surrounding text, remain under-explored in agent loops. In this paper, we report an empirical study organized into two experiments. Experiment 1 compares grep and vector retrieval on a 116-question sample from LongMemEval across four harnesses: a custom agent harness (Chronos) and three provider-native CLI harnesses (Claude Code, Codex, and Gemini CLI), under both inline tool results and file-based tool results that the model reads separately. Experiment 2 compares grep-only and vector-only retrieval while progressively mixing in additional unrelated conversation history, so that each query is embedded in more distracting material alongside the passages that matter. Across Chronos and the provider CLIs, grep consistently yields higher accuracy than vector retrieval in our comparisons, with inline grep exceeding inline vector for every harness-model pair we evaluate. At the same time, overall scores depend strongly on which harness and tool-calling style is used, even when the underlying conversation data are the same, with file-based delivery and provider CLI shells able to invert or erase the lexical advantage without any change to the corpus. These results highlight retrieval mechanics, harness orchestration, and delivery path as a single jointly evaluated system rather than independent design choices, motivating future work on hybrid retrieval policies, non-chat corpora, and broader vendor coverage to clarify when agents should reach for lexical compared to semantic search.


\bibliographystyle{ACM-Reference-Format}
\bibliography{references}

\clearpage 
\appendix

\section{Per-Category Accuracy}
\label{app:per-category}

Table~\ref{tab:category} reports accuracy by LongMemEval-S category for each inference model in the Chronos harness under grep-only retrieval, inline tool calling method, and the full haystack ($n{=}116$). Grader: GPT-4o.

\begin{table*}[t]
  \caption{Per-category accuracy (\%) on the 116-question subset for the Chronos harness (grep-only) at the full haystack, for each inference model. Grader: GPT-4o.}
  \label{tab:category}
  \centering
  \small
  \begin{tabular}{lcccccc}
    \toprule
    Model & KU & MS & SS-A & SS-P & SS-U & TR \\
    \midrule
    Claude Opus 4.6       & 94.4 & 83.9 & 100.0 & 100.0 & 87.5 & 87.1 \\
    Claude Haiku 4.5      & 83.3 & 71.0 & 100.0 & 85.7 & 87.5 & 87.1 \\
    GPT-5.4               & 77.8 & 74.2 & 92.3 & 85.7 & 93.8 & 67.7 \\
    Gemini 3.1 Pro        & 88.8 & 69.3 & 100.0 & 85.7 & 81.3 & 100.0 \\
    Gemini 3.1 Flash-Lite & 94.3 & 72.6 & 100.0 & 100.0 & 81.3 & 100.0 \\
    \bottomrule
  \end{tabular}
\end{table*}

\end{document}